%% file: main.tex
\documentclass[11pt,a4paper]{article}
\usepackage{authblk}
\usepackage[hyperref]{acl2019}
\usepackage{times}
\usepackage{latexsym}

\usepackage{floatrow}
\usepackage{paralist}
\usepackage{url}
\usepackage{multirow}
\usepackage[vlined,linesnumbered,resetcount]{algorithm2e}
\usepackage{amsmath,amssymb}

\usepackage{bbm}
\usepackage{rotating}
\usepackage{placeins}
\usepackage{subcaption}
\usepackage{graphicx}
\usepackage{mwe}
\usepackage{bm}
\usepackage{array}
\usepackage{xr}
\newcolumntype{C}[1]{>{\centering\let\newline\\\arraybackslash\hspace{0pt}}m{#1}}
\usepackage{pgfplotstable,colortbl}
\pgfplotsset{compat=1.12}

\usepackage{booktabs}
\usepackage{makecell}
\pgfplotsset{compat=1.12}

\pgfplotstableset{
  /color cells/min/.initial=0,
  /color cells/max/.initial=1,
  /color cells/textcolor/.initial=,
  color cells/.style={
    postproc cell content/.append code={%
          \pgfkeys{/color cells/.cd,#1}%
      \pgfkeysgetvalue{/pgfplots/table/@preprocessed cell content}\value%
      \ifx\value\empty%
      \else%
        \pgfmathfloatparsenumber{\value}\pgfmathfloattofixed{\pgfmathresult}%
        \let\value=\pgfmathresult%
        \pgfplotscolormapaccess%
            [\pgfkeysvalueof{/color cells/min}:\pgfkeysvalueof{/color cells/max}]%
            {\value}{\pgfkeysvalueof{/pgfplots/colormap name}}%
        \pgfkeysgetvalue{/pgfplots/table/@cell content}\typesetvalue%
        \pgfkeysgetvalue{/color cells/textcolor}\textcolorvalue%
        \toks0=\expandafter{\typesetvalue}%
        \edef\temp{\noexpand\pgfkeyssetvalue{/pgfplots/table/@cell content}{%
            \noexpand\cellcolor[rgb]{\pgfmathresult}%
            \noexpand\definecolor{mapped color}{rgb}{\pgfmathresult}%
            \ifx\textcolorvalue\empty\else\noexpand\color{\textcolorvalue}\fi%
            \the\toks0%
          }%
        }%
        \temp%
      \fi%
      }%
  }%
}%

\usepackage{tikz,enumerate}
\usepackage[framemethod=TikZ]{mdframed}
\usepackage{pgf}
\usetikzlibrary{automata}
\usetikzlibrary{shapes,positioning}
\usetikzlibrary{arrows}
\newcommand{\edge}[3][]{ %
  \foreach \x in {#2} { %
    \foreach \y in {#3} { %
      \path (\x) edge [->, >={triangle 45}, #1] (\y) ;%
    } ;
  } ;
}

\tikzstyle{latent} = [circle,fill=white,draw=black,inner sep=1pt,
minimum size=25pt, font=\fontsize{10}{10}\selectfont, node distance=1]
\tikzstyle{obs} = [latent,fill=gray!50]

\makeatletter
\renewcommand\paragraph{\indent \textbf}
\makeatother

\aclfinalcopy 
\def\aclpaperid{664} 

\title{Miss Tools and Mr Fruit: Emergent communication in agents learning about object affordances}

\author[1]{Diane Bouchacourt}

\author[1,2]{Marco Baroni}
\affil[1]{Facebook A.I. Research}
\affil[2]{ICREA}
\affil[ ]{\tt {\{dianeb,mbaroni\}@fb.com}}

\date{}

\begin{document}

\maketitle

\begin{abstract}
Recent research studies communication emergence in communities of deep network agents assigned a joint task,  hoping to gain insights on  human language evolution. We propose here a new task capturing crucial aspects of the human environment, such as natural object affordances, and of human conversation, such as full symmetry among the participants. By conducting a thorough pragmatic and semantic analysis of the emergent protocol, we show that the agents solve the shared task through genuine bilateral, referential communication. However, the agents develop multiple idiolects, which makes us conclude that full symmetry is not a sufficient condition for a common language to emerge.
\end{abstract}

\input{introduction}

\input{thegame}

\input{architectures}

\input{setup}

\input{results}

\input{related}

\input{discussion}

\section*{Acknowledgments}
We thank Rahma Chaabouni, Evgeny Kharitonov, Emmanuel Dupoux, Maxime Oquab and Jean-R{\'{e}}mi King for their useful discussions and insights. We thank David Lopez-Paz and Christina Heinze-Deml for their feedback on the causal influence of communication. We also thank Francisco Massa for his help on setting up the experiments.

\bibliography{marco,diane}
\bibliographystyle{acl_natbib}
\setcounter{table}{0}
\setcounter{figure}{0}
\renewcommand{\thetable}{A\arabic{table}}
\renewcommand{\thefigure}{A\arabic{figure}}
\input{appendices}

\end{document}

%% file: introduction.tex
\section{Introduction}
\label{sec:introduction}

The advent of powerful deep learning architectures has revived
research in simulations of language emergence among computational
agents that must communicate to accomplish a task
\cite[e.g.,][]{Jorge:etal:2016,Havrylov:Titov:2017,Kottur:etal:2017,Lazaridou:etal:2017,Lee:etal:2018,Choi:etal:2018,Evtimova:etal:2018,Lazaridou:etal:2018}. The
nature of the emergent communication code should provide insights on
questions such as to what extent comparable functional pressures could
have shaped human language, and whether deep learning
models can develop human-like linguistic skills. For such inquiries to
be meaningful, the designed setup should reflect as many aspects of
 human communication as possible. Moreover, appropriate tools
should be applied to the analysis of emergent communication, since, as
several recent studies have shown, agents might succeed at a task
without truly relying on their communicative channel, or by
means of \emph{ad-hoc} communication techniques overfitting their
environment
\cite{Kottur:etal:2017,Bouchacourt:Baroni:2018,Lowe:etal:2019}.

We contribute on both fronts. We introduce a game meeting many
desiderata for a natural communication environment. We further propose
a two-pronged analysis of emerging communication, at the
\emph{pragmatic} and \emph{semantic} levels. At the pragmatic level,
we study communicative acts from a functional perspective, measuring
whether the messages produced by an agent have an impact on the
subsequent behaviour of the other. At the semantic level, we decode
which aspects of the extra-linguistic context the agents refer to, and
how such reference acts differ between agents. Some of our conclusions
are positive. Not only do the agents solve the shared task, but genuine
bilateral communication helps them to reach higher reward. Moreover,
their referential acts are meaningful given the task, carrying the semantics of their input. However, we also
find that even perfectly symmetric agents converge to distinct
idiolects instead of developing a single, shared code.

\newcommand{\rulesep}{\unskip\ \vrule\ }

\begin{figure*}[tbp]
\centering
\begin{subfigure}[t]{0.32\textwidth}
\includegraphics[width=\textwidth]{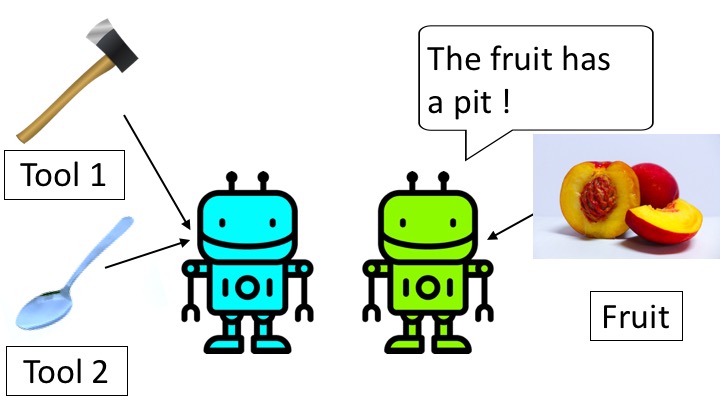}
\end{subfigure}
\rulesep
\begin{subfigure}[t]{0.32\textwidth}
\includegraphics[width=\textwidth]{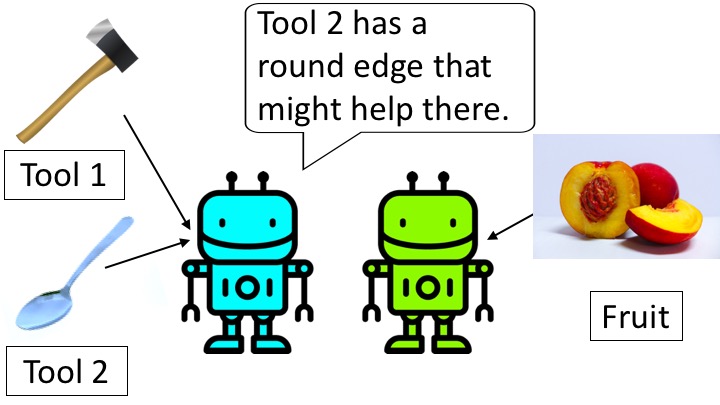}
\end{subfigure}
\rulesep
\begin{subfigure}[t]{0.32\textwidth}
\includegraphics[width=\textwidth]{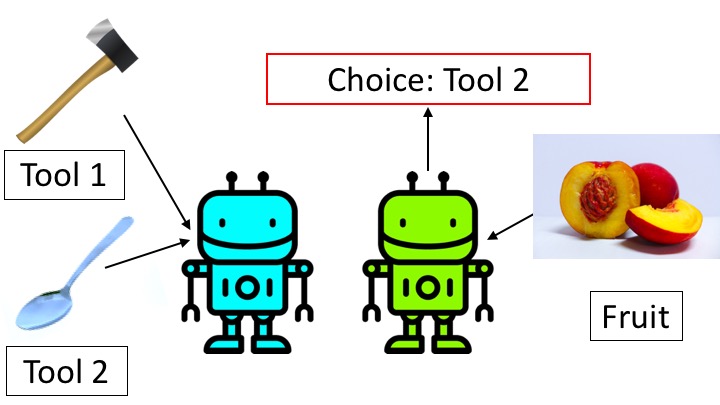}
\end{subfigure}
\caption{Our game. One agent receives a fruit, another two tools. Each
  agent sends a message in turn, until an agent ends the episode by
  choosing a tool. The agents are rewarded if the tool choice is
  optimal given the fruit.}
\label{fig::gameexemple}
\end{figure*}

%% file: thegame.tex
\section{The fruit and tools game}
\label{sec:thegame}

Our game, inspired by Tomasello's \shortcite{Tomasello:2014}
conjecture that the unique cognitive abilities of humans arose from
the requirements of cooperative interaction, is schematically
illustrated in Fig.~\ref{fig::gameexemple}. In each episode, a
randomly selected agent is presented with instances of two tools
(\emph{knife}, \emph{fork}, \emph{axe}\ldots), the other with a fruit
instance (\emph{apple}, \emph{pear}, \emph{plum}\ldots).
Tools and fruits are represented by property vectors (e.g., \emph{has
  a blade}, \emph{is small}), with each instance characterized by
values randomly varying around the category mean (e.g., an apple
instance might be smaller than another). An agent is randomly selected
to be the first to perform an action. The game then proceeds for an
arbitrary number of \emph{turns}. At each turn, one of the agents must
decide whether to pick one of the two tools and stop, or to
continue, in which case the message it utters is passed to the other
agent, and the game proceeds. Currently, for
ease of analysis, messages are single discrete symbols selected from a
vocabulary of size $10$, but extension to symbol sequences is
trivial (although it would of course complicate the analysis). As soon as an agent picks a tool, the game ends. The agents
receive a binary reward of 1 if they picked the better tool for the
fruit at hand, 0 otherwise. The best choice is computed by a utility
function that takes into account the interaction between tool and
fruit instance properties (e.g., as in Fig.~\ref{fig::gameexemple}, a
tool with a round edge might be particularly valuable if the fruit has
a pit). Utility is relative: given a peach, the axe is worse than the
spoon, but it would be the
better tool when the alternative is a hammer.\\
\indent Here are some desirable properties of our setup, as a
simplified simulation of human interactions. The agents are fully
symmetric and cannot specialize to a fixed role or turn-taking
scheme. The number of turns is open and determined by the agents. In
pure signaling/referential games \citep{Lewis:1969}, the aim is
successful communication itself. In our game, reward depends instead
on tool and fruit affordances. Optimal performance can only be
achieved by jointly reasoning about the properties of the tools and
how they relate to the fruit. Humans are rewarded when
they use language to solve problems of this sort, and not for
successful acts of reference \emph{per se}. Finally, as we use
commonsense descriptions of everyday objects to build our dataset (see
below), the distribution of their properties possesses the highly
skewed characteristics encountered everywhere in the human environment
\cite{Li:2002}. For example, if the majority of fruits requires to be cut, a knife is intrinsically more useful than a spoon. Note that the agents do not have any a priori knowledge of the tools utility. Yet, baseline agents are able to discover context-independent tool
affordances and already reach high performance. We believe that this
scenario, in which communication-transmitted information complements
knowledge that can be directly inferred by observing the world, is
more interesting than typical games in which
language is the only information carrier.\\
\paragraph{Game ingredients and utility} We picked 16 tool and 31 fruit
categories from \newcite{McRae:etal:2005} and
\newcite{Silberer:etal:2013}, who provide subject-elicited
property-based commonsense descriptions of objects, with some
extensions. We used $11$ fruit and $15$ tool features from these
databases to represent the categories. We rescaled the
elicitation-frequency-based property values provided in the norms to
lie in the $[0,1]$ range, and manually changed some counter-intuitive
values. An object instance is a property vector sampled from the
corresponding category as follows. For binary properties such as
\emph{has a pit}, we use Bernoulli sampling with $p$ equaling the
category value. For continuous properties such as \emph{is small}, we
sample uniformly from $[\mu - 0.1,\mu + 0.1]$, where $\mu$ is the
category value. We then devised a function returning an utility score
for any fruit-tool property vector pair. The function maps
properties to a reduced space of abstract functional features (such as
\emph{break} for tools, and \emph{hard} for fruits). Details are in
Supplementary Section \ref{suppsec:data}. For example, an apple with
\emph{is crunchy=0.7} value gets a high \emph{hard}  functional feature
score. A knife with \emph{has a blade=1} gets a high \emph{cut} score,
and therefore high utility for the \emph{hard}
apple. 
Some features, e.g., \emph{has a handle} for tools, have no impact on
utility. They only represent realistic aspects of objects and act as
noise. Our dataset with full category property vectors will be
publicly released along with code.\\
\begin{figure*}[t!]
\includegraphics[width=\textwidth]{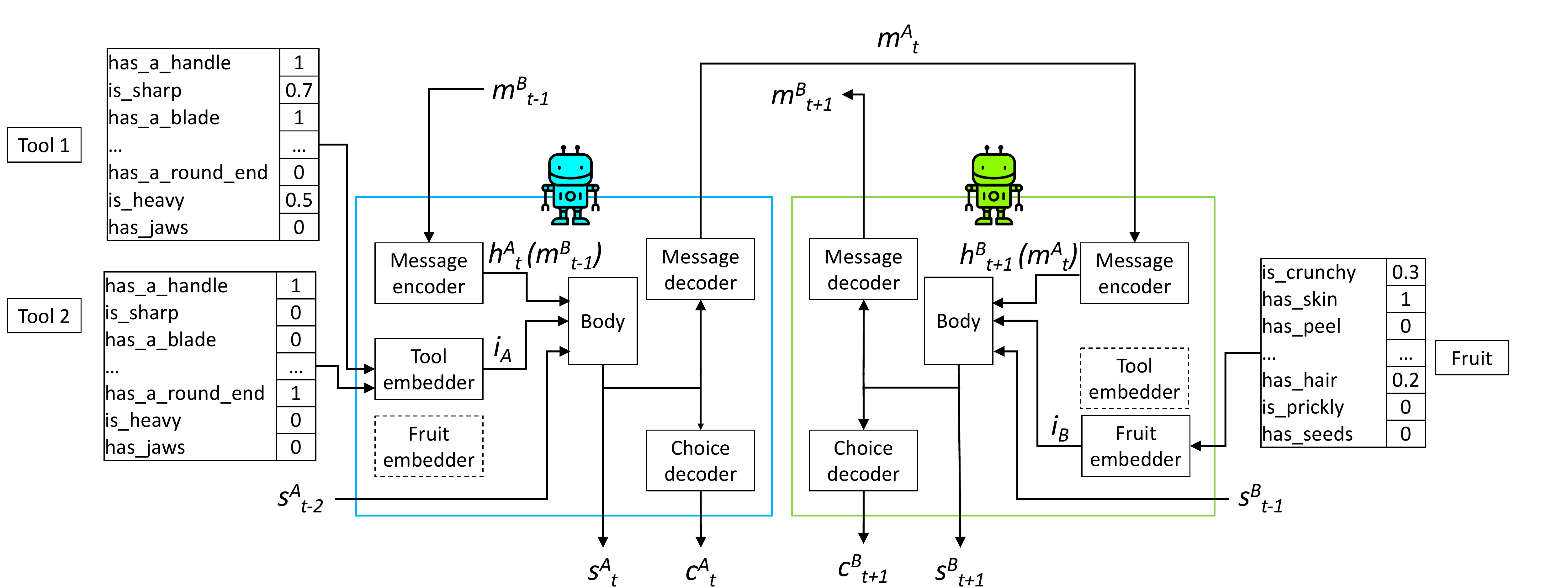}
\caption{Two turns of dialogue. Dashed boxes are not used in this game episode due to the agent roles (the blue/left agent is Tool Player, the green/right one is Fruit player). The flow is explained in detail in the text.}
\label{fig:oneround}
\end{figure*}
\paragraph{Datasets} We separate the $31$ fruit categories into three
sets: in-domain ($21$ categories), validation and transfer ($5$
categories each). The in-domain set is further split into train and
test partitions. We train agents on the in-domain train partition and
monitor convergence on the validation set. We report test performance
on the in-domain test partition and on the transfer set. For example,
the \emph{peach} category is in-domain, meaning that distinct peach
instances will be seen at training and in-domain testing time. The
\emph{nectarine} category is in the transfer set, so nectarine
instances will only be seen at test time. This scheme tests the
generalization abilities of the agents (that can generalize to new
fruits since they are all are described in the same feature space). We
generate $210,000$ in-domain training samples and $25,000$ samples for
the other sets, balanced across fruits and tools (that are common
across the sets).

%% file: architectures.tex
\paragraph{Game dynamics and agent architecture} At the beginning of a
game episode, two neural network agents A and B receive, randomly,
either a pair of tools $(tool_1,tool_2)$ (always sampled from
different categories) or a $fruit$. The agent receiving the tools
(respectively the fruit) will be Tool Player (respectively Fruit
Player) for the episode.\footnote{Agents must learn to recognize the
  assigned role.} The agents are also randomly assigned positions, and
the one in position 1 starts the game. Figure \ref{fig:oneround} shows
two turns of the game in which A (blue/left) is Tool Player and in
position 1. The first turn is indexed $t=0$, therefore A will act on
even turns, B on odd turns. At game opening, each agent passes its
input (tool pair or fruit) through a linear layer followed by a tanh
nonlinearity, resulting in embedding $i_A$ (resp.~$i_B$). Then, at
each turn $t$, an agent, for example A, receives the message
$m^B_{t-1}$ from agent B, and accesses its own previous internal state
$s^A_{t-2}$ (we refer to ``memory'' the addition of the agent's previous state). The message $m^B_{t-1}$ is processed by
a RNN, and the resulting hidden state $h^A_t(m^B_{t-1})$ is
concatenated with the agent previous internal state $s^A_{t-2}$ and
the input embedding $i_A$. The concatenated vector is fed to the Body
module, composed of a linear layer followed by tanh. The
output of the Body module is the new A state, $s^A_t$, fed to
Message and Choice decoders.

The Message decoder is an RNN with hidden state
initialized as $s^A_t$, and outputting a probability distribution
$p(m^A_t|s^A_t)$ over possible A messages. At training time, we
sample a message $m^A_t$; at test time we take
the most probable one. The Choice decoder is a linear
layer processing $s^A_t$ and outputting a softmax-normalized vector of
size $3$. The latter represents the probabilities $p(c^A_t|s^A_t)$
over A's possible choices: (i) $c^A_t=0$ to
continue playing, (ii) $c^A_t=1$ to choose tool $tool_1$ and stop (iii)
$c^A_t=2$ to choose tool $tool_2$ and stop. Again, we sample at training
and argmax at test time. If $c^A_t=0$, the game continues. B receives
 message $m^A_t$, its previous state $s^B_{t-1}$ and input embedding $i_B$, and it outputs the tuple ($m^B_{t+1}$, $c^B_{t+1}$, $s^B_{t+1}$) etc., until an agent stops the game, or the maximum number of turns $T_{\text{max}}=20$ is reached.

 When an agent stops by choosing a tool, for example $tool_1$, we
 compute the two utilities $U(tool_1,fruit)$ and $U(tool_2,fruit)$. If
 $U(tool_1,fruit)\ge U(tool_2,fruit)$, that is the best tool was chosen, shared reward is $R=1$. If
 $U(tool_1,fruit)<U(tool_2,fruit)$ or if the agents reach
 $T_{\text{max}}$ turns without choosing, $R=0$.\footnote{We also
   tried directly using raw or normalized scalar utilities as rewards,
   with similar performances.} During learning, the reward is
 back-propagated with Reinforce \citep{Williams:1992}. When the game
 starts at $t=0$, we feed the agent in position $1$ a fixed dummy
 message $m^0$, and the previous states of the agents $s^A_{t-2}$ and
 $s^B_{t-1}$ are initialized with fixed dummy $s^0$. In the no-memory
 ablation, previous internal states are always replaced by $s^0$. When we block
 communication, agent messages are replaced by $m^0$. Supplementary
 Section \ref{suppsec:imp_details} provides hyperparameter and training
 details.

%% file: setup.tex
\section{Measuring communication impact}
\label{sec:setup}

\paragraph{Message Effect} is computed
on single turns and uses causal theory \citep{Pearl:etal:2016} to
quantify how much what an agent utters impacts the other, compared to
the counterfactual scenario in which the speaking agent said something
else.

Consider message $m^A_t$ uttered by A at turn $t$. If
$c^A_t=0$ (that is, A continues the game), $m^A_t$ is processed by
 B, along with $s^B_{t-1}$ and $i^B$. At the next turn, B outputs
a choice $c^B_{t+1}$ and a message $m^B_{t+1}$ drawn from
$p(c^B_{t+1},m^B_{t+1}|s^B_{t+1})$. B's state $s^B_{t+1}$ is
deterministically determined by $m^A_t,c^A_t,s^B_{t-1},i^B$, so we can
equivalently write that $c^B_{t+1}$ and $m^B_{t+1}$ are drawn from
$p(c^B_{t+1},m^B_{t+1}|m^A_t,c^A_t,s^B_{t-1},i^B)$. Conditioning on
$c^A_t,s^B_{t-1},i^B$ ensures there are no confounders when we analyze
the influence from $m^A_t$ \citep{Pearl:etal:2016}. Supplementary Figure
\ref{fig:causalgraph} shows the causal graph supporting our assumptions. We
will not from here onwards write the conditioning on
$c^A_t,s^B_{t-1},i^B$ explicitly.

We define $z^B_{t+1}=(c^B_{t+1},m^B_{t+1})$. We want to estimate how much the message from A, $m^A_t$, influences the next-turn behaviour (choice and message) of B, $z^B_{t+1}$. 
We thus measure the discrepancy between the conditional distribution $p(z^B_{t+1}|m^A_t)$ and the marginal distribution $p(z^B_{t+1})$ not taking $m^A_t$ into account. However, we want to assess agent B's behaviour under \emph{other possible received messages} $m^A_t$. To do so, when we compute the marginal of agent B's $p(z^B_{t+1})$, we intervene on $m^A_t$ and draw the messages from the \emph{intervention distribution}. We define $\tilde{p}(z^B_{t+1})$, the marginal computed with counterfactual messages $m'^A_t$, as:
\begin{align}
\tilde{p}(z^B_{t+1})=\sum_{m^A_t}p(z^B_{t+1}|m'^A_t)\tilde{p}(m'^A_t)
\label{eq:margint}
\end{align}
where $\tilde{p}(m'^A_t)$ is the intervention distribution, different from $p(m^A_t|s^A_t)$. If at turn $t$, A continues the game, we define the \emph{Message Effect} ($\text{ME}$) from agent A's message $m^A_t$ on agent B's choice and message pair, $z^B_{t+1}$ as:
\begin{align}
&\text{ME}^{A\rightarrow B}_t=\text{KL}\Big(p(z^B_{t+1}|m^A_t)||\tilde{p}(z^B_{t+1})\Big)
\end{align}
where $\text{KL}$ is the Kullback-Leibler divergence, and $\tilde{p}(z^B_{t+1})$ is computed as in Eq.~\ref{eq:margint}. This allows us to measure how much the conditional distribution differs from the marginal. Algorithm \ref{algo:algoME} shows how we estimate $\text{ME}^{A\rightarrow B}_t$. In our experiments, we draw $K=10$ samples $z^B_{t+1,k}$, and use a uniform intervention distribution $\tilde{p}(m'^A_t)$ with $J=10$. This kind of counterfactual reasoning is explored in depth by \citet{Bottou:etal:2013}. \citet{Jaques:etal:2018} and \newcite{Lowe:etal:2019} present related
measures of causal impact based on the Mutual Information (MI) between
influencing and influenced agents. We discuss in Supplementary Section
\ref{suppsec:me} possible issues with the MI-based approach.

\newcommand{\nextnr}{\stepcounter{AlgoLine}\ShowLn}
\RestyleAlgo{boxed}
\SetAlgoCaptionSeparator{}
\LinesNotNumbered
\begin{algorithm}[t]
Given the message $m^A_t$ from agent A.\\
\nextnr Sample $K$ pairs $z^B_{t+1,k}\sim p(z^B_{t+1}|m^A_t)$.\label{step:sampB}\\
\nextnr Sample $J$ counterfactuals $m'^A_{t,j} \sim \tilde{p}(m'^A_t)$. \label{step:sampCount}\\
\nextnr For each $z^B_{t+1,k}$, do
\label{step:marg}
\vspace{-5pt}
\begin{align}
&\tilde{p}(z^B_{t+1,k})=\sum_{j=1}^{J}p(z^B_{t+1,k}|m'^A_{t,j})\tilde{p}(m'^A_{t,j}). \nonumber
\end{align}
\nextnr Return
\vspace{-5pt}
\begin{align}
\text{ME}^{A\rightarrow B}_t=\frac{1}{K}\sum_{k=1}^{K}\log\frac{p(z^B_{t+1,k}|m^A_t)}{\tilde{p}(z^B_{t+1,k})}. \nonumber
\end{align}
\caption{Computation of $\text{ME}^{A\rightarrow B}_t$.}
\label{algo:algoME}
\end{algorithm}

\paragraph{Bilateral communication} Intuitively, there has been a proper dialogue if, in the course of a conversation, each agent has said at least one thing that influenced the other. We operationalize this through our \emph{bilateral communication}
measure. This is a binary, per-game score, that is positive only if in
the game there has been at least one turn with significant message
effect \emph{in both directions}, i.e., $\exists~t,~t'~s.t.~\text{ME}^{A\rightarrow B}_t > \theta$ and $\text{ME}^{B \rightarrow A}_{t'} > \theta$. We set $\theta=0.1$.\footnote{We considered setting $\theta$ to (i) the average $\text{ME}$ returned by untrained agents, but this led to a threshold extremely close to $0$, and (ii) the average of the agents' $\text{ME}$ values, but this counterintuitively penalized pairs of agents with high overall communication influence.}

%% file: results.tex
\section{Results}

We first confirm that the agents succeed at the task, and
communication improves their performance. Second, we study their
\emph{pragmatics}, looking at how ablating communication
and memory affect their interaction. Finally, we try to interpret the \emph{semantics} of the agents' messages.

\input{performance_pragmatics}

\input{semantics}

%% file: performance_pragmatics.tex
\subsection{Performance and pragmatics}
\label{sec:prag}

We report mean and standard error of the mean (SEM) over successful training seeds.\footnote{That is, training seeds leading to final validation performance above $85\%$.} 
Each agent A or B can be either $F$ (Fruit Player) or $T$ (Tool Player) and in position $1$ or $2$, depending on the test game. We measure to what extent Tool Player influences  Fruit Player ($\text{ME}^{T\rightarrow F}$) and \emph{vice versa} ($\text{ME}^{F\rightarrow T}$). Similarly, we evaluate position impact by computing $\text{ME}^{1\rightarrow 2}$ and $\text{ME}^{2\rightarrow 1}$. We average ME values over messages sent during each test game, and report averages over test games. Note that we also intervene on the dummy initialization message used at $t=0$, which is received by the agent in position $1$. This impacts the value of $\text{ME}^{2\rightarrow 1}$. If the agent in position $1$ has learned to rely on the initialization message to understand that the game is beginning, an intervention on this message will have an influence we want to take into account.\footnote{Conversely, we  ignore the messages agents send when stopping the game, as they are never heard.} Similarly, in the no-communication ablation, when computing ME values, we replace the dummy fixed message the agents receive with a counterfactual. Finally, we emphasize that the computation of ME values does not interfere with game dynamics and does not affect performance.

\begin{table*}[t!]
  \begin{center}
\begin{tabular}{|c|c|cc|cc|}
  \hline
  ~& ~&\multicolumn{2}{c|}{\bfseries No communication} &\multicolumn{2}{c|}{\bfseries With communication}\\
\hline
~ & Metric & In & Transfer & In & Transfer \\
\hline
\multirow{ 7}{*}{\rotatebox[origin=c]{90}{\bfseries No memory}}&
Av. perf. (\%)& $84.83\pm0.09$& $84.0\pm0.11$&$96.9\pm0.32$& $94.5\pm0.37$\\
~&$ME^{F\rightarrow T}$& $0.133^*\pm0.01$& $0.14^*\pm0.01$&$5.0^*\pm0.39$& $5.0^*\pm0.36$\\
~&$ME^{T\rightarrow F}$& $0.05\pm0.02$& $0.030\pm0.01$&$3.9\pm0.38$& $3.3\pm0.30$\\
~&$ME^{1\rightarrow 2}$& $0.066\pm0.00$& $0.067\pm0.01$&$3.9\pm0.29$& $3.7\pm0.26$\\
~&$ME^{2\rightarrow 1}$& $0.12^*\pm0.02$& $0.10^*\pm0.01$&$5.0^*\pm0.38$& $4.7^*\pm0.33$\\
~&Bi. comm. (\%)& $1.4\pm0.31$& $1.3\pm0.40$&$\bm{88\pm2.49}$& $\bm{89\pm2.24}$\\
~&Av. conv. length& $0.508\pm0.01$& $0.52\pm0.01$&$2.16\pm0.08$& $2.21\pm0.10$\\
~&T chooses (\%)&$99.4\pm0.63$& $99.6\pm0.56$&$85\pm2.09$& $83\pm2.56$\\
\hline
\multirow{ 7}{*}{\rotatebox[origin=c]{90}{\bfseries With memory}}&
Av. perf. (\%)& $88.5\pm0.11$& $87.7\pm0.16$&$\bm{97.4\pm0.12}$& $\bm{95.3\pm0.16}$\\
~&$ME^{F\rightarrow T}$& $0.11^*\pm0.01$& $0.13^*\pm0.01$&$3.0^*\pm0.29$& $2.8^*\pm0.24$\\
~&$ME^{T\rightarrow F}$& $0.064\pm0.01$& $0.071\pm0.01$&$1.8\pm0.22$& $1.8\pm0.21$\\
~&$ME^{1\rightarrow 2}$& $0.085\pm0.01$& $0.10\pm0.01$&$2.4\pm0.29$& $2.3\pm0.22$\\
~&$ME^{2\rightarrow 1}$& $0.093\pm0.01$& $0.103\pm0.01$&$2.4\pm0.22$& $2.4\pm0.21$\\
~&Bi. comm. (\%)& $3.8\pm0.61$& $4.6\pm0.68$& $78\pm2.55$& $78\pm2.65$\\
~&Av. conv. length& $1.50\pm0.06$& $1.46\pm0.06$&$2.7\pm0.11$& $2.7\pm0.11$\\
~&T chooses (\%)&$87.3\pm1.34$& $85.8\pm1.48$&$81\pm2.94$& $81\pm3.00$\\
\hline
\end{tabular}
\caption{Test performance and pragmatic measures mean and SEM in different settings. ``Av.~perf.'' (average performance) denotes $\%$ of samples where best tool was chosen, ``Bi.~comm." denotes $\%$ of games with bilateral communication taking place. ``Av.~conv.~length" is average conversation length in turns. ``T~chooses" denotes $\%$ of games ended by Tool Player. Values of $\text{ME}$ with an asterisk $~^*$ are statistically significantly higher than their reverse (e.g. $\text{ME}^{F\rightarrow T}>\text{ME}^{T\rightarrow F}$). Best ``Av.~perf.'' and ``Bi.~comm." in bold.}
\label{tab:perftable}
\end{center}
\end{table*}

\paragraph{Both communication and memory help} Table \ref{tab:perftable} shows that enabling the agents to communicate greatly increases performance compared to the no-communication ablation, both with and without memory, despite the high baseline set by agents that learn about tool usefulness without communicating (see discussion below). Agents equipped with memory  perform better than their no-memory counterparts, but the gain in performance is smaller compared to the gain attained from communication.  The overall best performance is achieved with communication \emph{and} memory. We also see that the agents generalize well, with transfer-fruit performance almost matching that on in-domain fruits. Next, we analyze in detail the impact of each factor (communication, memory) on agent performance and strategies.\\
\paragraph{No-communication, no-memory} We start by looking at how the
game unfolds when communication and agent memory are ablated (top left
quadrant of Table \ref{tab:perftable}). Performance is largely above
chance ($\approx{}50\%$), because, as discussed in Section
\ref{sec:thegame}, some tools are intrinsically better on average
across fruits than others. 
Without
communication, the agents exploit this bias and learn a strategy where
(i) Fruit Player never picks the tool but always continues the
game and (ii) Tool Player picks the tool according to average tool
usefulness. Indeed, Tool
Player makes the choice in more than $99\%$ of the games. Conversation length is $0$ if Tool Player starts and $1$ if it is
the second agent, requiring the starting Fruit Player to pass its
turn.  Reassuringly, \text{ME} values are low, confirming the
reliability of this communication score, and indicating that
communication-deprived agents did not learn to rely on the fixed dummy
message (e.g., by using it as a constant bias). Still, we observe
that, across the consistently low values, Fruit Player appears to
affect Tool Player
significantly more than the reverse ($\text{ME}^{F\rightarrow T}>\text{ME}^{T\rightarrow F}$). This is generally observed in all configurations, and we believe it due to the fact that Tool Player takes charge of most of the reasoning in the game. We come back to this later in our analysis. We also observe that the second player impacts the first more than the reverse ($\text{ME}^{2\rightarrow 1} > \text{ME}^{1\rightarrow 2}$). We found this to be an artifact of the strategy adopted by the agents. In the games in which Tool Player starts and immediately stops the game, we can only compute ME for the Tool/position-1 agent, by intervening on the initialization. The resulting value, while tiny, is unlikely to be exactly $0$. In the games where Fruit Player starts and Tool Player stops at the second turn, we compute instead two tiny MEs, one per agent. Hence, the observed asymmetry. We verified this hypothesis by removing single-turn games: the influence of the second player on the first indeed disappears.\\
\paragraph{Impact of communication} The top quadrants of
Table \ref{tab:perftable} show that communication helps performance, despite the high baseline set by the
``average tool usefulness'' strategy.  Importantly, when communication
is added, we see a dramatic increase in the proportion of games
with bilateral communication, confirming that
improved performance is not due to an accidental effect of adding a new channel \cite{Lowe:etal:2019}. ME and average number of turns
also increase. Fruit Player is the more influential agent. This effect is not due to the artifact we found in the no-communication ablation, because almost all conversations, including those started by Tool Player, are longer than one turn, so we can compute  both $\text{ME}^{F\rightarrow T}$ and $\text{ME}^{T\rightarrow F}$. We believe the asymmetry to be due to the fact that  Tool Player is the agent that demands more information from the other, as it is the one that sees the tools, and that in the large majority of cases makes the final choice. Supplementary Table \ref{tab:perftable2} shows that the gap between the influence of the Fruit Player on the Tool player and its reverse is greater when the Fruit Player is in position $2$. This, then, explains $\text{ME}^{2\rightarrow 1} > \text{ME}^{1\rightarrow 2}$ as an epiphenomenon of Fruit Player being more influential.\\
%
\paragraph{Is memory ablation necessary for communication to matter?}
An important observation from previous research is that depriving at
least one agent of memory might be necessary to develop successful
multi-turn communication
\cite{Kottur:etal:2017,Cao:etal:2018,Evtimova:etal:2018}. This is
undesirable, as obviously language should not emerge simply as a
surrogate for memories of amnesiac agents. The performance and
communicative behaviours results in the bottom right quadrant of Table
\ref{tab:perftable} show that, in our game, genuine linguistic
interaction (as cued by ME and bilateral communication scores) is
present even when both agents are equipped with memory. It is
interesting however to study how adding memory affects the game
dynamics independently of communication. In the bottom left quadrant,
we see that memory leads to some task performance improvement for
communication-less agents. Manual inspection of example games reveals
that such agents are developing turn-based strategies.  For example,
Tool Player learns to continue the game at turn $t$ if $tool_1$ has a
round end. At $t+1$, Fruit Player can use the fact that Tool Player
continues at $t$ as information about relative tool roundness, and
either pick the appropriate one based on the fruit or continue to
gather more information. In a sense, agents learn to use the
possibility to stop or continue at each turn as a rudimentary
communication channel. Indeed, exchanges are on average longer when
memory is involved, and turn-based strategies appear even with
communication. In the latter case, agents rely on communication but also on turn-based
schemes, resulting in lower \text{ME} values and bilateral
communication compared to the no-memory ablation. Finally, the
respective positions of the agents in the conversation no longer
impact ME
($\text{ME}^{1\rightarrow 2}\approx{}\text{ME}^{2\rightarrow
  1}$). This might be because, with memory, the starting agent can
identify whether it is at turn $t=0$, where it almost always chooses
to continue the game to send and receive more information via
communication. Intervening on the dummy initialization message has a
lower influence, resulting in lower $\text{ME}^{2\rightarrow 1}$.

%% file: semantics.tex
\subsection{Conversation semantics}

Having ascertained that our agents are conducting bidirectional
conversations, we next try to decode what are the \emph{contents} of
such conversations. To do this, we train separate classifiers to predict, from the message exchanges in successful in-domain test game, what are Fruit, Tool 1, Tool 2 in the game.\footnote{We focus on
  the in-domain set as there are just 5 transfer fruit categories. We
  also tried predicting triples at once with a single classifier, that
  consistently reached above-baseline but very low accuracies.}
Consider for example a game in which fruit is \emph{apple} and tools 1
and 2 \emph{knife} and \emph{spoon}, respectively. If the
message-based classifiers are, say, able to successfully decode
\emph{apple} but not \emph{knife}/\emph{spoon}, this suggests that the
messages are about the fruit but not the tools. For each prediction
task, we train classifiers (i) on the whole conversation, i.e., both
agents' utterances (\emph{Both}), and (ii) on either Player's
utterances: Fruit (\emph{F}) or Tool only (\emph{T}). For comparison,
we also report accuracy of a baseline that makes guesses based on the
train category distribution (\emph{Stats}), which is stronger than
chance. We report  mean accuracy and SEM across successful training seeds. Supplementary Section \ref{suppsec:sem} provides further details
on classifier implementation and training.

\begin{table}[t]
\setlength{\tabcolsep}{2pt}
\begin{center}
\begin{tabular}{ |c|ccc| }
\hline
 Messages & Fruit & Tool 1 & Tool 2\\
\hline
Both& $37\pm1.70$& $31\pm1.21$& $24\pm1.07$\\
F& $37\pm1.75$& $23.3\pm0.66$& $16.7\pm0.51$\\
T& $14.1\pm0.79$& $32\pm1.17$& $25\pm1.04$\\
\hline
Stats. (\%)& $5.786\pm0.00$& $8.76\pm0.01$& $7.682\pm0.01$\\
\hline
\end{tabular}
\end{center}
\caption{Semantic classifier \% accuracy mean and SEM over successful training seeds.}
\label{tab:sem}
\end{table}
%

The first row of Table \ref{tab:sem} shows that the conversation as a whole carries information about any object. The second and third show that the agents are mostly conveying information about their respective objects (which is very reasonable), but also, to a lesser extent, but still well above baseline-level, about the other agent's input. This latter observation is intriguing. Further work should ascertain if it is an
artifact of fruit-tool correlations, or pointing in the direction of
more interesting linguistic phenomena (e.g., asking
``questions''). The asymmetry between Tool 1 and 2 would also deserve
further study, but importantly the agents are clearly referring to
both tools, showing they are not adopting entirely degenerate
strategies.\footnote{We experiment with single symbol messages (and multi-turn conversation) but using longer messages we could potentially witness interesting phenomena such as the emergence of compositionality. We leave this exploration for future work.}

We tentatively conclude that the agents did develop the expected
semantics, both being able to refer to all objects in the games. Did they
however developed \emph{shared} conventions to refer to them, as in
human language? This would not be an unreasonable expectation, since
the agents are symmetric and learn to play both roles and in both
positions. Following up on the idea of ``self-play'' of
\citet{Grasser:etal:2019}, after a pair of agents A and B are trained,
we replace at test time agent B's embedders and modules with those in
A, that is, we let one agent play with a copy of itself. If A and B
are speaking the same language, this should not affect test
performance. Instead, we find that with self-play average game
performance drops down to $67\%$ and $65 \%$ in in-domain and transfer test
sets, respectively. This suggests that the agents developed their own
idiolects. The fact that performance is still above chance could be
due to the fact that the latter are at least partially exchangeable,
or simply to the fact that agents can still do reasonably well by
relying on knowledge of average tool usefulness (self-play performance
is below that of the communication-less agents in Table
\ref{tab:perftable}). To decide between these interpretations, we
trained the semantic classifier on conversations where A is the Fruit
Player and B the Tool Player, testing on conversations about the same
inputs, but where the roles are inverted. The performance drops down
to the levels of the Stats baseline (Supplementary Table
\ref{tab:invres}), supporting the conclusion that non-random
performance is due to knowledge acquired by the agents independently
of communication, and not partial similarity among their codes.

%% file: related.tex
\section{Related work}

\paragraph{Games} Among the long history of early works that model language evolution between agents \cite[e.g.][]{Steels:2003b,Brighton:etal:2003}, \newcite{Reitter:Lebiere:2011} simulate human language evolution with a Pictionary type task. Most recently, with the advent of neural network architectures, literature focuses on simple
referential games with a sender sending a single message to a
receiver, and reward depending directly on communication success
\cite[e.g.,][]{Lazaridou:etal:2017,Havrylov:Titov:2017,Lazaridou:etal:2018}. \newcite{Evtimova:etal:2018}
extend the referential game presenting the sender and receiver with
referent views in different modalities, and allowing multiple
message rounds. Still, reward is given directly for referential
success, and the roles and turns of the agents are
fixed. \newcite{Das:etal:2017b} generalize Lewis' signaling game \citep{Lewis:1969} and propose a cooperative image guessing game between two agents, a question bot and an answer bot. They find that grounded language emerges without supervision. \newcite{Cao:etal:2018} \citep[expanding on][]{Lewis:etal:2017}
propose a setup where two agents see the same set of items, and each
is provided with arbitrary, episode-specific utility functions for the
object. The agents must converge in multi-turn conversation to a
decision about how to split the items. The fundamental novelty of our
game with respect to theirs is that our rewards depend on consistent,
realistic commonsense knowledge that is stable across episodes
(hammers are good to break hard-shell fruits,
etc.). \citet{Mordatch:Abbeel:2018} \citep[see also][]{Lowe:etal:2017}
study emergent communication among multiple ($>2$) agents pursuing
their respective goals in a maze. In their setup, fully symmetric
agents are encouraged to use flexible, multi-turn communication as a
problem-solving tool. However, the independent complexities of
navigation make the environment somewhat cumbersome if the aim is to study
emergent communication.

\paragraph{Communication analysis} Relatively few papers have focused
specifically on the analysis of the emergent communication
protocol. Among the ones more closely related to our line of inquiry,
\newcite{Kottur:etal:2017} analyze a multi-turn signaling game. One
important result is that, in their game, the agents only develop a sensible code if the sender is deprived of memory across
turns. \newcite{Evtimova:etal:2018} study the dynamics of agent
confidence and informativeness as a conversation progresses. \newcite{Cao:etal:2018} train probe classifiers to predict, from the messages, each agent utility function and the decided split of items. Most directly related to our pragmatic analysis, \newcite{Lowe:etal:2019},
who focus on simple matrix communication games, introduce the notions
of positive signaling (an agent sends messages that are related to its
state) and positive listening (an agent's behaviour is influenced by
the message it receives). They show that positive signaling does not
entail positive listening, and commonly used metrics might not
necessarily detect the presence of one or the other. We build on their
work, by focusing on the importance of mutual positive listening in
communication (our ``bilateral communication'' measure). We
further refine the causal approach to measuring influence they
introduce. \citet{Jaques:etal:2018} also use the notion of causal
influence, both directly as a term in the agent cost function, and to
analyze their behaviour.

%% file: discussion.tex
\section{Discussion}
\label{sec:discussion}

We introduced a more challenging and arguably natural game to study
emergent communication in deep network agents. Our experiments show
that these agents do develop genuine communication even when
\begin{inparaenum}[(i)]
\item successful communication \emph{per se} is not directly rewarded;
\item the observable environment already contains stable, reliable
  information helping to solve the task (object affordances); and
\item the agents are not artificially forced to rely on communication
  by erasing their memory.
\end{inparaenum} The linguistic exchanges of the agents are not only
leading to significantly better task performance, but can be properly
pragmatically characterized as dialogues, in the sense that the
behaviour of each agent is affected by what the other agent
says. Moreover, they use language, at least in part, to denote the
objects in their environment, showing primitive hallmarks of a
referential semantics.

We also find, however, that agent pairs trained together in fully
symmetrical conditions develop their own idiolects, such that an agent
won't (fully) understand itself in self play. As convergence to a
shared code is another basic property of human language, in
future research we will explore ways to make it emerge. First, we
note that \citet{Grasser:etal:2019}, who study a simple signaling
game, similarly conclude that training single pairs of agents does not
lead to the emergence of a common language, which requires diffusion
in larger communities. We intend to verify if a similar trend emerges
if we extend our game to larger agent groups. Conversely, equipping
the agents with a feedback loop in which they also receive their own
messages as input might encourage shared codes across speaker and
listener roles.

In the current paper, we limited ourselves to one-symbol messages,
facilitating analysis but greatly reducing the spectrum of potentially
emergent linguistic phenomena to study. Another important direction
for future work is thus to endow agents with the possibility of
producing, at each turn, a sequence of symbols, and analyze how this
affects conversation dynamics and the communication protocol. Finally,
having shown that agents succeed in our setup, we intend to test them
with larger, more challenging datasets, possibly involving more
realistic perceptual input.

%% file: appendices.tex
\section{Data and utility computation}
\label{suppsec:data}
This section provides additional details on the dataset we use and the utility function we employ to compute the utilities between fruits and tools. Note that we refer to fruits for conciseness, but some vegetables, such as \emph{carrot} and \emph{potato}, are included.


There are $11$ fruits features: \emph{is crunchy, has skin, has peel, is small, has rough skin, has a pit, has milk, has a shell, has hair, is prickly, has seeds} and $15$ tools features: \emph{has a handle, is sharp, has a blade, has a head, is small, has a sheath, has prongs, is loud, is serrated, has handles, has blades, has a round end, is adorned with feathers, is heavy, has jaws}. Note that, when we sample instances of each category as explained in Section \ref{sec:thegame} of the main paper, %
 features are sampled independently. We filter out, however, nonsensical combinations. For example, the features \emph{has prongs}, \emph{has a blade} and \emph{has blades} are treated as pairwise mutually exclusive.

In order to compute the utility for a pair ($tool$, $fruit$), we use three mapping matrices. The mapping matrix $M_T \in \mathbb{R}^{15\times6}$ (Table \ref{tab:mt}) maps from the space of tool features to a space of more general functional features: \emph{(cut, spear, lift, break, peel, pit remover)},
and similarly $M_F \in \mathbb{R}^{11\times6}$ (Table \ref{tab:mf}) maps from the space of fruits features to a space of functional features: \emph{(hard, pit, shell, pick, peel, empty inside)}. Finally, the matrix $M \in \mathbb{R}^{6\times6}$ (Table \ref{tab:m}) maps the two abstract functional spaces of features together. For example, if an axe sample is described by the vector $t_a \in \mathbb{R}^{1\times15}$ and a nectarine sample is the vector $f_n \in \mathbb{R}^{1\times11}$, the utility is computed as $U(t_a,f_n)=(f_n M_F) M' (t_a M_T)'$ where $'$ denotes transpose. We always add a value of $0.01$ to avoid zero utilities. Therefore we can compute the utility of any combination of (possibly new) fruits and tools, as long as it can be described in the corresponding functional representational space. Note that in our case we have the same number of abstract functional features for fruits and tools ($6$), but they need not be the same. In other words, $M$ need not be a square matrix.

Given the values in the mapping matrices, $5$ of the tools features have no impact on the utility computation since they do not affect the scores of the functional tool features (they have only zeros in the mapping matrix $M_T$). These are: \emph{has a handle, is sharp, has a sheath, is loud, has handles, is adorned with feathers}. Such features only represent realistic aspects of objects and act as noise.

\begin{table*}
  \setlength{\tabcolsep}{2pt}
  \begin{center}
\pgfplotstabletypeset[col sep=comma,
/color cells/max=1,
/color cells/min=-1,
/color cells/textcolor=black,
columns/Cut/.style={color cells},
columns/Spear/.style={color cells},
columns/Lift/.style={color cells},
columns/Break/.style={color cells},
columns/Peel/.style={color cells},
columns/Pit Remover/.style={color cells},
columns/Tools Feature/.style={string type, column type=c|},
/pgfplots/colormap/violet,
every head row/.style={before row=\toprule,after row=\midrule},
every last row/.style={after row=\bottomrule},
]{
Tools Feature,Cut,Spear,Lift,Break,Peel,Pit Remover
has a handle,0,0,0,0,0,0
is sharp,0,0,0,0,0,0
has a blade,1,0.5,0,0,1,0
has a head,0,0,0,1,0,0
is small,0,0,0,0,0,0.25
has a sheath,0,0,0,0,0,0
has prongs,0.5,1,0.25,0,0.25,0
is loud,0,0,0,0,0,0
is serrated,0.5,0,0,0,0,0
has handles,0,0,0,0,0,0
has blades,1,0.5,0,0,0.5,0
has a round end,0.25,0,1,0,0,1
is adorned with feathers,0,0,0,0,0,0
is heavy,0,0,0,0.5,0,0
has jaws,0,0,1,0,0,0.5
}
\caption{$M_{T}$. Rows are dataset tool features, columns are functional tool features.}
\label{tab:mt}
\end{center}
\end{table*}

\begin{table*}
    \setlength{\tabcolsep}{2pt}
  \begin{center}
\pgfplotstabletypeset[col sep=comma,
/color cells/max=1,
/color cells/min=-1,
/color cells/textcolor=black,
columns/Hard/.style={color cells},
columns/Shell/.style={color cells},
columns/Pick/.style={color cells},
columns/Pit/.style={color cells},
columns/Peel/.style={color cells},
columns/Empty inside/.style={color cells},
columns/Fruits Feature/.style={string type, column type=c|},
every head row/.style={before row=\toprule,after row=\midrule},
every last row/.style={after row=\bottomrule},
/pgfplots/colormap/violet,
]{
Fruits Feature,Hard,Pit,Shell,Pick,Peel,Empty inside
is crunchy,1,0,0,0,0,0
has skin,0,0,0,0,1,0
has peel,0,0,0,0,1,0
is small,0,0,0,1,0,0
has rough skin,0,0,0.5,0,0,0
has a pit,0,1,0,0,0,0
has milk,0,0,0,0,0,1
has a shell,0,0,1,0,0,0
has hair,0,0,0,0,0.5,0
is prickly,0,0,0,0,0.5,0
has seeds,0,0,0,0,0,1
}
\caption{$M_{F}$. Rows are dataset fruit features, columns are functional fruit features.}
\label{tab:mf}
\end{center}
\end{table*}

\begin{table*}
    \setlength{\tabcolsep}{2pt}
    \begin{center}
\pgfplotstabletypeset[col sep=comma,
/color cells/max=1,
/color cells/min=-1,
/color cells/textcolor=black,
columns/Hard/.style={color cells},
columns/Shell/.style={color cells},
columns/Pick/.style={color cells},
columns/Pit/.style={color cells},
columns/Peel/.style={color cells},
columns/Empty inside/.style={color cells},
columns/Feature/.style={string type, column type=c|, column name={}},
every head row/.style={before row=\toprule,after row=\midrule},
every last row/.style={after row=\bottomrule},
/pgfplots/colormap/violet,
]{
Feature,Hard,Pit,Shell,Pick,Peel,Empty inside
Cut,1,0,0.5,0,0.5,0
Spear,0,0,0,1,0,0
Lift,0,0,0,0.5,0,1
Break,0.5,0,1,0,0,0
Peel,0,0,0,0,1,0
Pit Remover,0,1,0,0,0,0
}
\caption{$M$. Rows are functional tool features, columns are functional fruit features.}
\label{tab:m}
\end{center}
\end{table*}

\section{Implementation details}
\label{suppsec:imp_details}

\subsection{Training and architecture hyperparameters}

We update the parameters with RMSProp \citep{Tieleman:Hinton:2012} with a learning rate of $0.001$ and the rest of the parameters left to their Pytorch default value. We use a scalar reward baseline $b$ to reduce variance, learned with Mean Square Error such that $1+b$ matches the average reward. We clip all gradients at 0.1. For the Message encoder and decoder modules, we embed input and output symbols with dimensionality $50$ and then use a RNN with $100$ hidden dimensions. The Fruit embedder linear transformation is of output size $100$, the Tool embedder is of size $50$. The Body module is of size $100$. We train the agents with batches of $128$ games for a total of $1$ million batches. We validate on $12$ batches of $100$ games, for a total of $1200$ validation games, and similarly for testing. 

\subsection{Test procedure details}

The computation of the $\text{ME}$ values involves random sampling in steps \ref{step:sampB} and \ref{step:sampCount} of Algorithm \ref{algo:algoME} so we test using $20$ testing seeds. For each successful training seeds, we compute the average ME over the test seeds, and report the mean and standard error of the mean (SEM) of the average ME. Given two trained agents A and B, there are $C=4$ possibles configurations at test time:
\begin{enumerate}
\item A is Fruit Player/position 1 and B is Tool Player/position 2
\item A is Fruit Player/position 2 and B is Tool Player/position 1
\item A is Tool Player/position 1 and B is Fruit Player/position 2
\item A is Tool Player/position 2 and B is Fruit Player/position 1
\end{enumerate}

We balance the number of test games in each configuration: we use $3$ batches of $100$ test games in each configuration, resulting in $12$ batches for a total of $1200$ test games. The ME value in each configuration $c$ is the average ME over the number of batches in this configuration ($3$ in our case). We then average the ME in each configuration over the four possible configurations to obtain $ME^{1\rightarrow 2}$, $ME^{F\rightarrow T}$ and their reverse.

\section{Message effect metric}
\label{suppsec:me}
\subsection{Causal graph and assumptions}

\begin{figure}[h!]
  \begin{center}
\begin{tikzpicture}[every node/.style={minimum size=15cm,inner sep=0,outer sep=0}]
  \node[obs]    (sb)  {$s^B_{t-1}$}; %
  \node[latent,below=1cm of sb]    (mb1)  {$m^B_{t-1}$}; %
  \node[latent,right=0.5cm of mb1]    (cb1)  {$c^B_{t-1}$}; %

  \node[latent,below=of mb1]    (sa)  {$s^A_t$}; %
  \node[obs,fill={rgb:orange,1;yellow,2;pink,5},below=1cm of sa]    (ma)  {$m^A_t$}; %
  \node[obs,right=0.5cm of ma]    (ca)  {$c^A_t$}; %
  \node[latent,left=2cm of sa]    (ia)  {$i^A$}; %
  \node[latent,above left=1.6cm of mb1]    (sa1)  {$s^A_{t-2}$}; %

  \node[obs,right=2cm of sb]    (ib)  {$i^B$}; %
  \node[latent,below=of ma] (sb2)      {$s^B_{t+1}$} ; %
  \node[latent, below=1cm of sb2]      (mb)      {$m^B_{t+1}$} ; %
  \node[latent, right=0.5cm of mb]       (cb)      {$c^B_{t+1}$} ;

  \node[latent,below=of mb] (sa2)      {$s^A_{t+2}$} ; %

  \edge {sb} {mb1} ; %
  \edge {sb} {cb1} ; %
  \edge {mb1} {sa} ; %
  \edge {cb1} {sa} ; %
    \edge {ia} {sa} ; %
    \edge {sa1} {sa} ; %
  \edge {sa} {ma} ; %
  \edge {sa} {ca} ; %
  \edge {sa} {ma} ; %
  \edge {sa} {ca} ; %
  \edge {ca} {sb2} ; %
  \edge {ma} {sb2} ;
    \edge {ib} {sb.east} ; %
  \draw[->,>={triangle 45}] (ib) to[bend left=45] (sb2.east);
  \draw[->,>={triangle 45}] (ia) to[bend right=20] (sa2);
  \draw[->,>={triangle 45}] (sb) to[bend left=70] (sb2);
  \edge {sb2} {mb} ; %
  \edge {sb2} {cb} ;
  \edge {mb} {sa2} ; %
  \edge {cb} {sa2} ; %
  \edge {ia} {sa1} ; %
  \draw[->,>={triangle 45}] (sa) to[bend right=25] (sa2);
\end{tikzpicture}
\caption{Causal graph considered when we compute $\text{ME}^{A\rightarrow B}_t$. The orange node $m^A_t$ is the variable we intervene on. Shaded nodes represent the variables we condition on.}
\label{fig:causalgraph}
\end{center}
\end{figure}
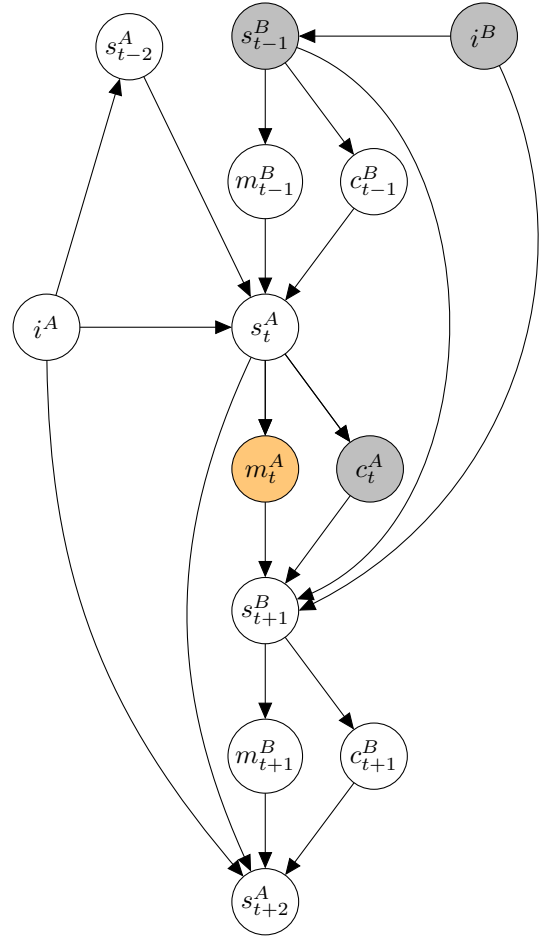

Figure \ref{fig:causalgraph} shows the causal graph we consider when computing
$\text{ME}^{A\rightarrow B}_t$. We write all variables that should be considered at this turn $t$. Conditioning on $c^A_t,s^B_{t-1},i^B$ blocks any backdoor paths when we compute the causal influence of $m^A_t$ on $c^B_{t+1},m^B_{t+1}$ \citep{Pearl:etal:2016}. Moreover, the path between $m^A_t$ and $c^B_{t+1},m^B_{t+1}$ through $i^A$ is blocked at the collider node $s^A_{t+2}$. Therefore, we ensure there is no confounder when we compute the influence of $m^A_t$ on $c^B_{t+1},m^B_{t+1}$. As in \citet{Jaques:etal:2018}, we have knowledge of the inputs to the model and the distributions of the variables we consider. Therefore we do not need to perform abduction to update probabilities of unobserved exogenous variables that may alter the causal relations in our model \citep{Pearl:etal:2016}.

We denote agent B's choice and message pair at turn $t+1$ as $z^B_{t+1}=(c^B_{t+1},m^B_{t+1})$. We explain in Section \ref{sec:setup} of the main paper that we compare (i) the conditional distribution $p(z^B_{t+1}|m^A_t)$ and (ii) the marginal distribution $p(z^B_{t+1})$ which does not take $m^A_t$ into account. We intervene on $m^A_t$, and draw counterfactual messages not from agent A but from another distribution over $m^A_t$, the \emph{intervention distribution}. We define $\tilde{p}(z^B_{t+1})$, the marginal computed with counterfactual messages $m'^A_t$, as:
\begin{align}
\tilde{p}(z^B_{t+1})=\sum_{m^A_t}p(z^B_{t+1}|m'^A_t)\tilde{p}(m'^A_t).\label{eqsupp:marg1}
\end{align}
where $\tilde{p}(m'^A_t)$ is the \emph{intervention distribution}. In our experiments we take a uniform intervention distribution. Importantly, $\tilde{p}(m'^A_t)$ is different from the \emph{observational distribution} distribution $p(m^A_t|s^A_t)$ that agent A actually defines over the messages. Contrarily to \citet{Bottou:etal:2013} by feeding the counterfactuals messages to agent B, we access $p(z^B_{t+1}|m'^A_t)$ and need not estimate it from empirical data.

\subsection{Difference with Mutual Information}

\citet{Jaques:etal:2018} train agents to have impact on other agents
by maximizing the causal influence of their actions. They show their
definition of influence to relate to the Mutual Information (MI)
between influencing and influenced agents. \newcite{Lowe:etal:2019} also
define a causal influence metric based on MI.  MI computation
requires counterfactuals drawn from the influencing agent
distribution, and not from an intervention one. In our setting, this
means drawing counterfactuals from agent A's distribution
$p(m^A_t|s^A_t)$, and not $\tilde{p}(m'^A_t)$, in step
\ref{step:sampCount} of Algorithm \ref{algo:algoME} (main paper).

There is an issue with employing MI and drawing counterfactuals from the influencing agent's distribution, e.g., $p(m^A_t|s^A_t)$, that is particularly pressing if message distributions are very skewed (as it is the case with our agents below). Consider a simple setting where agents A and B utter a message that has two possible values $u$ and $v$. A is the influencing agent and B is the influencee. Suppose that the dynamics are such that A almost always says $u$, and B always replies with the message it received. Most of the exchanges we would sample would be: ``A says $u$, B replies $u$". The MI estimate would then be very low, and one might erroneously conclude that B is not influenced by A. Indeed when distributions are very peaky as in this example, it would require very large samples to witness rare events, such as ```A says $v$, B replies $v$". \newcite{Lowe:etal:2019} ensure that all possible messages from the influencing agent are considered. This is computationally expensive when the message space is large. Moreover, the resulting MI can still be small as each message's contribution is weighted by its probability under the influencing agent. By using a uniform intervention distribution, we ensure that B in the current example would receive $v$ and therefore reply $v$ in half the exchanges, easily detecting the influence of A on B.

\section{Additional results: performance and pragmatics}
\label{suppsec:addres}

Table \ref{tab:perftable2} reports a more detailed view of the ME in Table \ref{tab:perftable} of the main paper. $1T/2F$ denotes games where the Tool Player is in first position and the Fruit Player is in second position, and $1F/2T$ denotes games where the Fruit Player in first position and Tool Player in second. This table shows that in the no-memory, with-communication setting (top right quadrant), the difference between the influence of the Fruit Player on the Tool player and its reverse is greater when the Fruit Player is in position $2$. On in-domain data, the Fruit Player has a stronger influence on the Tool Player only when in position $2$. This explains the effect $\text{ME}^{2\rightarrow 1} > \text{ME}^{1\rightarrow 2}$ we mention in Section \ref{sec:prag} in the main paper. We also observe that in the no-memory, no-communication setting (top left quadrant), when the Tool Player is in position 1, $ME^{1\rightarrow 2}~1T/2F \approx 0$. This relates to the artifact we describe in the main paper: in that case the Tool Player stops the game at $t=0$, leaving no room for the Fruit Player to be influenced.

\begin{table*}[h!]
  \begin{center}
\begin{tabular}{|c|c|cc|cc|}
  \hline
  ~& ~&\multicolumn{2}{c|}{\bfseries No communication} &\multicolumn{2}{c|}{\bfseries With communication}\\
\hline
~ & Metric & In & Transfer & In & Transfer \\
\hline
\multirow{ 7}{*}{\rotatebox[origin=c]{90}{\bfseries No memory}}&
Av. perf. (\%)& $84.83\pm0.09$& $84.0\pm0.11$&$96.9\pm0.32$& $94.5\pm0.37$\\
~&$ME^{F\rightarrow T}$& $0.133^*\pm0.01$& $0.14^*\pm0.01$&$5.0^*\pm0.39$& $5.0^*\pm0.36$\\
~&$ME^{T\rightarrow F}$& $0.05\pm0.02$& $0.030\pm0.01$&$3.9\pm0.38$& $3.3\pm0.30$\\
~&$ME^{1\rightarrow 2}$& $0.066\pm0.00$& $0.067\pm0.01$&$3.9\pm0.29$& $3.7\pm0.26$\\
~&$ME^{2\rightarrow 1}$& $0.12^*\pm0.02$& $0.10^*\pm0.01$&$5.0^*\pm0.38$& $4.7^*\pm0.33$\\
~&$ME^{1\rightarrow 2}~1T/2F$& $0.000001\pm0.00$& $0.000001\pm0.00$&$3.7\pm0.46$& $3.0\pm0.34$\\
~&$ME^{2\rightarrow 1}~1T/2F$& $0.13^*\pm0.01$& $0.15^*\pm0.01$&$5.8^*\pm0.50$& $5.7^*\pm0.47$\\
~&$ME^{1\rightarrow 2}~1F/2T$& $0.133\pm0.01$& $0.13^*\pm0.01$&$4.2\pm0.34$& $4.4^*\pm0.34$\\
~&$ME^{2\rightarrow 1}~1F/2T$& $0.10\pm0.03$& $0.06\pm0.02$&$4.2\pm0.39$& $3.6\pm0.29$\\
\hline
\multirow{ 7}{*}{\rotatebox[origin=c]{90}{\bfseries With memory}}&
Av. perf. (\%)& $88.5\pm0.11$& $87.7\pm0.16$&$97.4\pm0.12$& $95.3\pm0.16$\\
~&$ME^{F\rightarrow T}$& $0.11^*\pm0.01$& $0.13^*\pm0.01$&$3.0^*\pm0.29$& $2.8^*\pm0.24$\\
~&$ME^{T\rightarrow F}$& $0.064\pm0.01$& $0.071\pm0.01$&$1.8\pm0.22$& $1.8\pm0.21$\\
~&$ME^{1\rightarrow 2}$& $0.085\pm0.01$& $0.10\pm0.01$&$2.4\pm0.29$& $2.3\pm0.22$\\
~&$ME^{2\rightarrow 1}$& $0.093\pm0.01$& $0.103\pm0.01$&$2.4\pm0.22$& $2.4\pm0.21$\\
~&$ME^{1\rightarrow 2}~1T/2F$& $0.063\pm0.01$& $0.064\pm0.01$&$1.8\pm0.24$& $1.8\pm0.23$\\
~&$ME^{2\rightarrow 1}~1T/2F$& $0.12^*\pm0.02$& $0.13^*\pm0.02$&$2.9^*\pm0.25$& $2.9^*\pm0.25$\\
~&$ME^{1\rightarrow 2}~1F/2T$& $0.106^*\pm0.01$& $0.13^*\pm0.02$&$3.1^*\pm0.35$& $2.8^*\pm0.25$\\
~&$ME^{2\rightarrow 1}~1F/2T$& $0.065\pm0.01$& $0.077\pm0.01$&$1.8\pm0.21$& $1.9\pm0.20$\\
\hline
\end{tabular}
\caption{Detailed ME values (compare to Table \ref{tab:perftable} in main paper). $1T/2F$ denotes games where the Tool Player is in first position and the Fruit Player is in second position, and $1F/2T$ denotes games where the Fruit Player in first position and Tool Player in second.}
\label{tab:perftable2}
\end{center}
\end{table*}

\begin{table*}[t!]
\begin{center}
  \small{
\begin{tabular}{ |c|ccc| }
\hline
 Utterances & Fruit & Tool 1 & Tool 2\\
\hline
Both, A is F & $42\pm2.21$& $32\pm2.04$& $27\pm1.33$\\
Both, B is F & $44\pm2.00$& $28\pm1.58$& $28\pm1.69$\\
\makecell{Both, Train A is F / Test B is F}& $6.8\pm0.61$& $11\pm1.16$& $8.8\pm0.68$\\
\makecell{Both, Train B is F / Test A is F} & $5.9\pm0.53$& $10\pm1.05$& $8.8\pm0.62$\\
\hline
Stats A is F & $6.4\pm0.27$& $8.9\pm0.42$& $8.2\pm0.38$\\
Stats B is F& $6.4\pm0.15$& $9.1\pm0.61$& $9.0\pm0.75$\\
\hline
\end{tabular}
}
\end{center}
\caption{Semantic classifier \% accuracy in inverted-roles setup}
\label{tab:invres}
\end{table*}

\section{Details on the semantics classifier}
\label{suppsec:sem}

\subsection{Classifier training and hyperparameters}

Our classifier consists of an Embedding table of size $50$ which maps the agents' discrete utterances to a continuous space, then uses a RNN with a hidden size of $100$ to map the entire embedded conversation into a hidden state. The hidden state is then fed to a linear classifier that predicts a score for each class, and the number of classes depends on the prediction task (e.g. $31$ classes when the task is to predict the fruit). We consider the setting with communication and with memory. From successful test in-domain conversations, we create train/validation/test partitions for the classifier. We ensure that each fruit is in the train set, and each tool in either of the two positions. We use $20$ different seeds for initializing the classifier dataset partitioning into train/validation/test. For each successful training seed, we compute the average accuracy over these $20$ test initialization seeds, and report the classifier accuracy mean and standard error of the mean (SEM) over the successful training seeds.

The agents were trained with symmetrical roles and random starting agent, but we generate  conversations with fixed roles and positions, so that all conversations follow the same pattern (for example: agent A always starts and agent A is always the Fruit Player).

\subsection{Inverted-roles experiment}

Table \ref{tab:invres} shows the results of the inverted-roles experiment: e.g., we train the classifier on conversations where A is Fruit Player and B is Tool Player, and test on conversations about the same inputs, but where the roles are inverted, that is, B is Fruit Player and A is Tool Player. The performance drops compared to testing on conversations where the roles are not inverted. For this experiment, we consider only the conversations that have at least one utterance from each agent (conversation length $\ge 2$) in order to remove the potential confounding effect of conversation length.

%% file: main.bbl
\begin{thebibliography}{29}
\expandafter\ifx\csname natexlab\endcsname\relax\def\natexlab#1{#1}\fi

\bibitem[{Bottou et~al.(2013)Bottou, Peters, {Qui\~{n}onero-Candela}, Charles,
  Chickering, Portugaly, Ray, Simard, and Snelson}]{Bottou:etal:2013}
L\'{e}on Bottou, Jonas Peters, Joaquin {Qui\~{n}onero-Candela}, Denis~X.
  Charles, D.~Max Chickering, Elon Portugaly, Dipankar Ray, Patrice Simard, and
  Ed~Snelson. 2013.
\newblock Counterfactual reasoning and learning systems: The example of
  computational advertising.
\newblock \emph{Journal of Machine Learning Research}, 14:3207--3260.

\bibitem[{Bouchacourt and Baroni(2018)}]{Bouchacourt:Baroni:2018}
Diane Bouchacourt and Marco Baroni. 2018.
\newblock How agents see things: On visual representations in an emergent
  language game.
\newblock In \emph{Proceedings of EMNLP}, pages 981--985, Brussels, Belgium.

\bibitem[{Brighton et~al.(2003)Brighton, Kirby, and Smith}]{Brighton:etal:2003}
Henry Brighton, Simon Kirby, and Kenneth Smith. 2003.
\newblock \emph{Situated cognition and the role of multi-agent models in
  explaining language structure.}, volume 2636, pages 88--109. Springer-Verlag
  GmbH.

\bibitem[{Cao et~al.(2018)Cao, Lazaridou, Lanctot, Leibo, Tuyls, and
  Clark}]{Cao:etal:2018}
Kris Cao, Angeliki Lazaridou, Marc Lanctot, Joel~Z. Leibo, Karl Tuyls, and
  Stephen Clark. 2018.
\newblock Emergent communication through negotiation.
\newblock In \emph{Proceedings of ICLR}.

\bibitem[{Choi et~al.(2018)Choi, Lazaridou, and {de Freitas}}]{Choi:etal:2018}
Edward Choi, Angeliki Lazaridou, and Nando {de Freitas}. 2018.
\newblock Compositional obverter communication learning from raw visual input.
\newblock In \emph{Proceedings of ICLR Conference Track}, Vancouver, Canada.
\newblock Published online:
  \url{https://openreview.net/group?id=ICLR.cc/2018/Conference}.

\bibitem[{Das et~al.(2017)Das, Kottur, Moura, Lee, and Batra}]{Das:etal:2017b}
Abhishek Das, Satwik Kottur, Jos{\'e} M.~F. Moura, Stefan Lee, and Dhruv Batra.
  2017.
\newblock Learning cooperative visual dialog agents with deep reinforcement
  learning.
\newblock In \emph{2017 IEEE International Conference on Computer Vision
  (ICCV)}.

\bibitem[{Evtimova et~al.(2018)Evtimova, Drozdov, Kiela, and
  Cho}]{Evtimova:etal:2018}
Katrina Evtimova, Andrew Drozdov, Douwe Kiela, and Kyunghyun Cho. 2018.
\newblock Emergent communication in a multi-modal, multi-step referential game.
\newblock In \emph{Proceedings of ICLR Conference Track}, Vancouver, Canada.
\newblock Published online:
  \url{https://openreview.net/group?id=ICLR.cc/2018/Conference}.

\bibitem[{Graesser et~al.(2019)Graesser, Cho, and Kiela}]{Grasser:etal:2019}
Laura Graesser, Kyunghyun Cho, and Douwe Kiela. 2019.
\newblock \href {https://arxiv.org/abs/1901.08706} {Emergent linguistic
  phenomena in multi-agent communication games}.
\newblock \emph{CoRR}.

\bibitem[{Havrylov and Titov(2017)}]{Havrylov:Titov:2017}
Serhii Havrylov and Ivan Titov. 2017.
\newblock Emergence of language with multi-agent games: Learning to communicate
  with sequences of symbols.
\newblock In \emph{Proceedings of NIPS}, pages 2149--2159, Long Beach, CA, USA.

\bibitem[{Jaques et~al.(2018)Jaques, Lazaridou, Hughes, G{\"{u}}l{\c{c}}ehre,
  Ortega, Strouse, Leibo, and de~Freitas}]{Jaques:etal:2018}
Natasha Jaques, Angeliki Lazaridou, Edward Hughes, {\c{C}}aglar
  G{\"{u}}l{\c{c}}ehre, Pedro~A. Ortega, DJ~Strouse, Joel~Z. Leibo, and Nando
  de~Freitas. 2018.
\newblock \href {http://arxiv.org/abs/1810.08647} {Intrinsic social motivation
  via causal influence in multi-agent {RL}}.
\newblock \emph{CoRR}, abs/1810.08647.

\bibitem[{Jorge et~al.(2016)Jorge, K{\aa}geb{\"a}ck, and
  Gustavsson}]{Jorge:etal:2016}
Emilio Jorge, Mikael K{\aa}geb{\"a}ck, and Emil Gustavsson. 2016.
\newblock Learning to play {Guess Who?}~and inventing a grounded language as a
  consequence.
\newblock In \emph{Proceedings of the NIPS Deep Reinforcement Learning
  Workshop}, Barcelona, Spain.
\newblock Published online:
  \url{https://sites.google.com/site/deeprlnips2016/}.

\bibitem[{Kottur et~al.(2017)Kottur, Moura, Lee, and Batra}]{Kottur:etal:2017}
Satwik Kottur, Jos{\'e} Moura, Stefan Lee, and Dhruv Batra. 2017.
\newblock Natural language does not emerge `naturally' in multi-agent dialog.
\newblock In \emph{Proceedings of EMNLP}, pages 2962--2967, Copenhagen,
  Denmark.

\bibitem[{Lazaridou et~al.(2018)Lazaridou, Hermann, Tuyls, and
  Clark}]{Lazaridou:etal:2018}
Angeliki Lazaridou, {Karl Moritz} Hermann, Karl Tuyls, and Stephen Clark. 2018.
\newblock Emergence of linguistic communication from referential games with
  symbolic and pixel input.
\newblock In \emph{Proceedings of ICLR Conference Track}, Vancouver, Canada.
\newblock Published online:
  \url{https://openreview.net/group?id=ICLR.cc/2018/Conference}.

\bibitem[{Lazaridou et~al.(2017)Lazaridou, Peysakhovich, and
  Baroni}]{Lazaridou:etal:2017}
Angeliki Lazaridou, Alexander Peysakhovich, and Marco Baroni. 2017.
\newblock Multi-agent cooperation and the emergence of (natural) language.
\newblock In \emph{Proceedings of ICLR Conference Track}, Toulon, France.
\newblock Published online:
  \url{https://openreview.net/group?id=ICLR.cc/2017/conference}.

\bibitem[{Lee et~al.(2017)Lee, Heo, and Zhang}]{Lee:etal:2018}
Sang-Woo Lee, Yu-Jung Heo, and Byoung-Tak Zhang. 2017.
\newblock Answerer in questioner's mind for goal-oriented visual dialogue.
\newblock \url{https://arxiv.org/abs/1802.03881}.

\bibitem[{Lewis(1969)}]{Lewis:1969}
David Lewis. 1969.
\newblock \emph{Convention}.
\newblock Harvard University Press, Cambridge, MA.

\bibitem[{Lewis et~al.(2017)Lewis, Yarats, Dauphin, Parikh, and
  Batra}]{Lewis:etal:2017}
Mike Lewis, Denis Yarats, Yann Dauphin, Devi Parikh, and Dhruv Batra. 2017.
\newblock Deal or no deal? {End-to-end} learning of negotiation dialogues.
\newblock In \emph{Proceedings of EMNLP}, pages 2443--2453, Copenhagen,
  Denmark.

\bibitem[{Li(2002)}]{Li:2002}
Wentian Li. 2002.
\newblock Zipf's law everywhere.
\newblock \emph{Glottometrics}, 5:14--21.

\bibitem[{Lowe et~al.(2019)Lowe, Foerster, Boureau, Pineau, and
  Dauphin}]{Lowe:etal:2019}
Ryan Lowe, Jakob Foerster, {Y-Lan} Boureau, Joelle Pineau, and Yann Dauphin.
  2019.
\newblock Measuring emergent communication is tricky.
\newblock In \emph{Proceedings of AAMAS}, Montreal, Canada.
\newblock {I}n press.

\bibitem[{Lowe et~al.(2017)Lowe, Wu, Tamar, Harb, Abbeel, and
  Mordatch}]{Lowe:etal:2017}
Ryan Lowe, Yi~Wu, Aviv Tamar, Jean Harb, Pieter Abbeel, and Igor Mordatch.
  2017.
\newblock Multi-agent actor-critic for mixed cooperative-competitive
  environments.
\newblock In \emph{{NIPS}}, pages 6382--6393.

\bibitem[{McRae et~al.(2005)McRae, Cree, Seidenberg, and
  McNorgan}]{McRae:etal:2005}
Ken McRae, George Cree, Mark Seidenberg, and Chris McNorgan. 2005.
\newblock Semantic feature production norms for a large set of living and
  nonliving things.
\newblock \emph{Behavior Research Methods}, 37(4):547--559.

\bibitem[{Mordatch and Abbeel(2018)}]{Mordatch:Abbeel:2018}
Igor Mordatch and Pieter Abbeel. 2018.
\newblock Emergence of grounded compositional language in multi-agent
  populations.
\newblock In \emph{{AAAI}}, pages 1495--1502. {AAAI} Press.

\bibitem[{Pearl et~al.(2016)Pearl, Glymour, and Jewell}]{Pearl:etal:2016}
Judea Pearl, Madelyn Glymour, and Nicholas Jewell. 2016.
\newblock \emph{Causal Inference in Statistics: A Primer}.
\newblock John Wiley \& Sons.

\bibitem[{Reitter and Lebiere(2011)}]{Reitter:Lebiere:2011}
David Reitter and Christian Lebiere. 2011.
\newblock How groups develop a specialized domain vocabulary: A cognitive
  multi-agent model.
\newblock \emph{Cognitive Systems Research}, 12:175--185.

\bibitem[{Silberer et~al.(2013)Silberer, Ferrari, and
  Lapata}]{Silberer:etal:2013}
Carina Silberer, Vittorio Ferrari, and Mirella Lapata. 2013.
\newblock Models of semantic representation with visual attributes.
\newblock In \emph{Proceedings of ACL}, pages 572--582, Sofia, Bulgaria.

\bibitem[{Steels(2003)}]{Steels:2003b}
Luc Steels. 2003.
\newblock Evolving grounded communication for robots.
\newblock \emph{Trends in cognitive sciences}, 7:308--312.

\bibitem[{Tieleman and Hinton(2012)}]{Tieleman:Hinton:2012}
Tijmen Tieleman and Geoffrey Hinton. 2012.
\newblock Lecture 6.5---rmsprop: Divide the gradient by a running average of
  its recent magnitude.
\newblock COURSERA: Neural Networks for Machine Learning.

\bibitem[{Tomasello(2014)}]{Tomasello:2014}
Michael Tomasello. 2014.
\newblock \emph{A Natural History of Human Thinking}.
\newblock Harvard University Press, Cambridge, MA.

\bibitem[{Williams(1992)}]{Williams:1992}
Ronald Williams. 1992.
\newblock Simple statistical gradient-following algorithms for connectionist
  reinforcement learning.
\newblock \emph{Machine learning}, 8(3-4):229--256.

\end{thebibliography}
